\documentclass{article}
\usepackage[preprint]{neurips_2022}
\usepackage[latin1]{inputenc}
\usepackage{graphicx}
\usepackage{amsmath}
\usepackage{amsfonts}
\usepackage{amssymb}
\usepackage{makeidx}
\usepackage{setspace}
\usepackage{geometry}
\usepackage{color}
\usepackage{hyperref}
\usepackage{natbib}
\usepackage{epigraph} 
\usepackage[table]{xcolor}
\title{Can deep learning match the efficiency of human visual long-term memory in storing object details?}
\author{%
  Emin Orhan \\
  New York University\\
  \texttt{aeminorhan@gmail.com}
  }
\date{}
\begin{document}

\maketitle

\begin{abstract}
Humans have a remarkably large capacity to store detailed visual information in long-term memory even after a single exposure, as demonstrated by classic experiments in psychology. For example, Standing (1973) showed that humans could recognize with high accuracy thousands of pictures that they had seen only once a few days prior to a recognition test. In deep learning, the primary mode of incorporating new information into a model is through gradient descent in the model's parameter space. This paper asks whether deep learning via gradient descent can match the efficiency of human visual long-term memory to incorporate new information in a rigorous, head-to-head, quantitative comparison. We answer this in the negative: even in the best case, models learning via gradient descent require approximately 10 exposures to the same visual materials in order to reach a recognition memory performance humans achieve after only a single exposure. Prior knowledge induced via pretraining and bigger model sizes improve performance, but these improvements are not very visible after a single exposure (it takes a few exposures for the improvements to become apparent), suggesting that simply scaling up the pretraining data size or model size might not be a feasible strategy to reach human-level memory efficiency.
\end{abstract}

\section{Introduction}

\epigraph{We, at one glance, can perceive three glasses on a table; Funes, all the leaves and tendrils and fruit that make up a grape vine. He knew by heart the focus of the southern clouds at dawn on the 30th of April, 1882, and could compare them in his memory with the mottled streaks on a book in Spanish binding he had only seen once and with the outlines of the foam raised by an oar in the Rio Negro the night before the Quebracho uprising. These memories were not simple ones; each visual image was linked to muscular sensations, thermal sensations, etc. He could reconstruct all his dreams, all his half-dreams. Two or three times he had reconstructed a whole day; he never hesitated, but each reconstruction had required a whole day. He told me: ``I alone have more memories than all mankind has probably had since the world has been the world.'' And again: ``My dreams are like you people's waking hours.'' And again, toward dawn: ``My memory, sir, is like a garbage heap.'' A circle drawn on a blackboard, a right triangle, a lozenge---all these are forms we can fully and intuitively grasp; Ireneo could do the same with the stormy mane of a pony, with a herd of cattle on a hill, with the changing fire and its innumerable ashes, with the many faces of a dead man throughout a long wake. I don't know how many stars he could see in the sky.}{--- \textit{Funes the Memorious}, Jorge Luis Borges}

Borges' famous short story, \textit{Funes the Memorious}, describes a fictional character, Ireneo Funes, who has an extraordinary capacity to remember things and events \citep{borges1962}. Funes' exceptional feats of memory, as recounted by Borges, inspire a sense of awe and fascination in us partly because they make us viscerally aware of the inferiority of our own capacity to remember in comparison. 

The memory capacity of an average human being is, of course, nowhere near as impressive as that of Funes, yet experiments repeatedly suggest that it may still be surprisingly large and that we may often be subjectively underestimating our own capacity to remember things \citep{shepard1967,standing1973,hollingworth2004}. For example, in a classic study, Lionel Standing showed that humans could recognize with high accuracy 10000 pictures they were shown only once a few days prior to a recognition test \citep{standing1973}.~In a more recent follow-up study, \citet{brady2008} showed that these long-term visual memories may be remarkably detailed, fine-grained memories.

How would our current machine learning models fare in their ability to incorporate and retain new visual information in a head-to-head comparison with humans? Would they already perform at super-human levels, a bit like real-world artificial versions of Ireneo Funes, or would they fall significantly short of the efficiency of human memory to incorporate new visual information? In current deep learning practice, the primary mode of incorporating new information into a model is through gradient descent in the model's parameter space (other less standard ways of incorporating information into a deep learning model are discussed in the Discussion section below). In this paper, we ask if deep learning via gradient descent can match the efficiency of human long-term memory to incorporate new visual information in a rigorous, head-to-head, quantitative comparison.

An answer to this question would be highly informative for a few reasons: (1) if current deep learning models turn out to be inferior to humans, we can aim for human-level memory efficiency as a feasible performance target for our models (similar to aiming for human-level Go playing, human-level machine translation, or human-level speech recognition); (2) if we find that deep learning models can in principle match human memory efficiency but only with much greater sample size requirements (\textit{e.g.} only after pretraining them with ultra large datasets), this can motivate us to improve the sample efficiency of our models relative to humans; (3) in the case of either (1) or (2), a better understanding and appreciation of how humans achieve high memory efficiency with low sample complexity can give us ideas or hints about how to improve our models in these respects.

Our results suggest that deep learning via gradient descent, as currently practiced, cannot match human memory efficiency when subjected to the same visual recognition memory experiment as in \citet{brady2008}. We find that even in the best case, deep learning models require roughly 10 exposures to the same visual stimuli in order to reach a recognition memory performance that humans achieve after a single exposure. Prior knowledge induced through pretraining and larger model sizes improve the model performance, however these improvements are usually not apparent after a single exposure, requiring at least a few exposures to begin to show their effects. This suggests that simply scaling up the pretraining dataset size or the model size may not be a feasible strategy to reach human-level memory efficiency, pointing instead to the importance of algorithmic improvements to achieve this goal. We discuss the implications of these findings in the context of earlier studies both in deep learning and in human memory research. All code, models, stimuli, and simulation results related to this work can be found at the following public repository: \url{https://github.com/eminorhan/igpt-memory}.

\section{Methods}
\subsection{Experimental setup}
Our main goal in this paper is to replicate the visual recognition memory experiments in \citet{brady2008} \textit{in silico} by subjecting deep learning models to the same experimental stimuli and conditions. 

\begin{figure}
  \centering
    \includegraphics[width=1.0\textwidth, trim=0mm 0mm 0mm 0mm, clip]{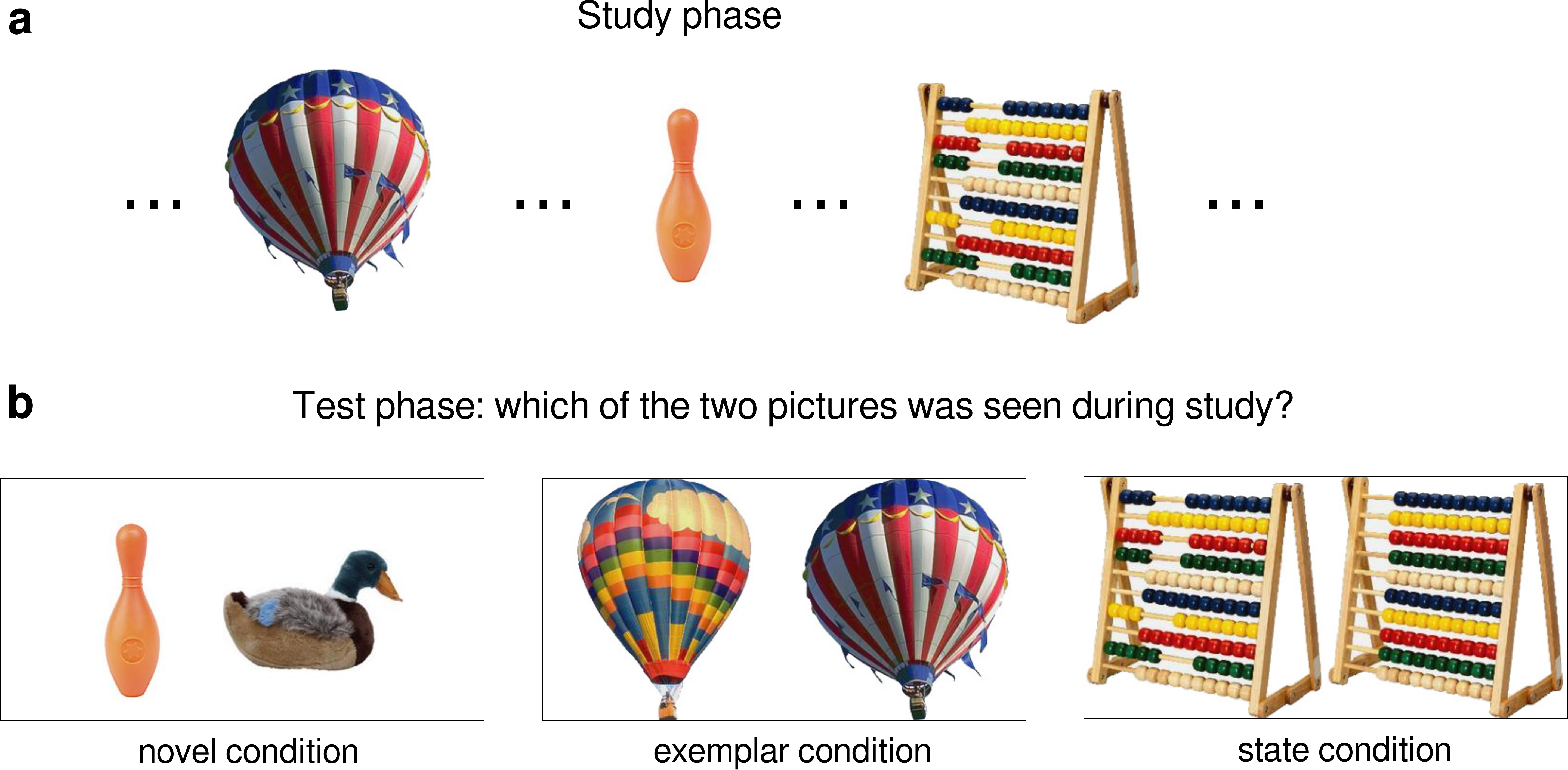}
  \caption{A schematic illustration of the basic design of the visual recognition memory experiment in \citet{brady2008}. (a) During the study phase, participants were sequentially shown 2500 pictures of real world objects. (b) After that, a recognition test followed, where in each trial a study item was paired with an unseen picture (foil) and participants were asked to indicate which of the two pictures was presented during the study phase. Different types of foils were used in different trials in order to probe the granularity of the participants' memory for the study items.}
  \label{brady_schema_fig}
\end{figure}

In these experiments, human subjects were first shown 2500 pictures of real world objects (study stimuli; see Figure~\ref{brady_schema_fig}a). Each picture was shown only once for 3 seconds during the study phase. After this study session, which took over 5 hours to complete, a test session followed in which in each trial, a randomly chosen study picture was paired with a new, unseen picture (foil) and subjects were asked to indicate which of the two pictures was seen during the study session. To probe the resolution of subjects' memory for the study stimuli, three different types of foil were used in the test trials: (i) \textit{novel}, where the foil came from a different category than all the study stimuli presented during the experiment, hence this condition only required a coarse conceptual representation of the study item in memory; (ii) \textit{exemplar}, where the foil came from the same category as the study item, but was a different exemplar, hence this condition presumably required a more detailed representation of the study item in memory in order to distinguish it from the foil; (iii) \textit{state}, where the foil depicted the same object as the study item but in a different state than was seen during the study phase, this condition thus presumably required the most detailed representation of the study item in memory, since the subject had to remember the particular state in which the study object was observed during the study phase. An example each of the three types of test trials is shown in Figure~\ref{brady_schema_fig}b. Each of these conditions consisted of 100 trials in the experiment, for a total of 300 test trials.

\subsection{Objective}
The recognition memory experiment described in the previous subsection is most naturally modeled as a probabilistic model learning task, where during the study phase, the model learns a probability density function, $p_{\theta}(\mathbf{x})$, that tries to maximize the likelihood of the study items: 
\begin{equation}
    \max_{\theta} \prod_{s \in S} p_{\theta}(\mathbf{x}_s)
\end{equation}
where $S$ denotes the set of study items. Equivalently, one can minimize the negative log-likelihood of the study items (which is the approach we will follow in this work):
\begin{equation}
    \min_{\theta} \mathcal{L}(\theta) \equiv \sum_{s \in S} -\log p_{\theta}(\mathbf{x}_s)
\end{equation}
It is crucial to note that because of the relatively small size of the study set (2500 images), given enough passes through the study set (as is commonly done in standard machine learning problems), a sufficiently large capacity model can successfully fit or ``memorize'' the study items through gradient descent, achieving a near-zero loss on the study set and a perfect accuracy on the test trials. Therefore, the key comparison in this case is between human performance in the experiments of \citet{brady2008} and the model performance after \textbf{\textit{only a single pass through the study set}}, \textit{i.e.} after only a single exposure to each study item (just like in the human experiments). In standard machine learning terminology, this corresponds to a single \textit{epoch} of training over the study set. Note that we do allow batch updates to the model during the study phase, hence we do not restrict ourselves to a strictly \textit{online} learning setting (we found that batch updates generally work better than online updates). Note also that this setting is different from the common one-shot (or few-shot) learning problem in machine learning, where the ``shot'' usually refers to the number of examples (or sample size), and not to the number of exposures to the examples.\footnote{One can formalize the notion of the \textit{number of exposures} to an image as the number of forward-props of the image through the model.}

For the test trials, we calculate the likelihoods of both the study item $s$ and the foil $f$ under the learned probabilistic model and the one that returns a higher likelihood is chosen as the model's response for that trial: $c=s$ if $p_{\theta}(\mathbf{x}_s)>p_{\theta}(\mathbf{x}_f)$, otherwise $c=f$ (where $c$ denotes the model's choice).

\subsection{Models}
\textbf{Model class:} We implement the probabilistic model, $p_{\theta}(\mathbf{x})$, with Image GPT (iGPT) models trained autoregressively \citep{chen2020}. To be able to feasibly model the images, we first resize the original images used by \citet{brady2008}, which are 256$\times$256 pixels, to 64$\times$64. As in \citet{chen2020}, we further eliminate the color channels (RGB) by running $k$-means on the pixels of the study set in the three-dimensional color space with a dictionary size of 512. Thus, the final preprocessed images, $\mathbf{x}$, are 64$\times$64 pixels in size where each pixel takes on a discrete value among 512 possible values. We refer the reader to \citet{chen2020} for further details about the iGPT model.

\textbf{Model size:} For the main experiments, we use iGPT models with 24 layers, 8 attention heads, and an embedding size of 512 (equivalent in size to the iGPT-S model in \citet{chen2020} with $\sim$77M parameters). To investigate the effect of model size on recognition memory performance, later on we also consider roughly 4$\times$ smaller models with 6 layers, 2 attention heads, and an embedding size of 512. We call this smaller model iGPT-mini.

\textbf{Prior knowledge:} To investigate the effect of prior knowledge on visual recognition memory performance, we consider both models trained from scratch on the study set (\textit{tabula rasa} models) and models pretrained on a large natural image dataset. We consider two such datasets for instilling prior knowledge about the visual world into the model: ImageNet-1K, which consists of $\sim$1.28M natural images of real-world objects from 1000 different categories \citep{russakovsky2015}, and SAYCam, a dataset of longitudinal headcam videos recorded from the perspective of three babies early in their development, \textit{i.e.} roughly from 6 to 32 months \citep{sullivan2020}. Further details about pretraining as well as samples generated from the pretrained iGPT models can be found in the Appendix. 

Our main motivation for using SAYCam is that, compared to ImageNet, it is a more representative sample of the visual experiences of humans (especially young humans), hence arguably it provides a better opportunity to conduct a fair comparison between human subjects and deep learning models. That being said, it should still be kept in mind that even the combined data from all three babies in SAYCam, approaching $\sim$500 hours total of natural video, is about two orders of magnitude smaller in size than what a typical sighted human being would experience by the age of 10. In our experiments with SAYCam, we use data from all three babies. We temporally subsample the SAYCam videos at a rate of 0.5 fps (1 frame every two seconds), resulting in a dataset with $\sim$0.86M frames in total.

\section{Results}
\subsection{Prior knowledge helps, but even the best models need $10\times$ more exposures than humans in order to reach human-level recognition memory performance}

\begin{figure}
  \centering
    \includegraphics[width=1.0\textwidth, trim=0mm 0mm 0mm 0mm, clip]{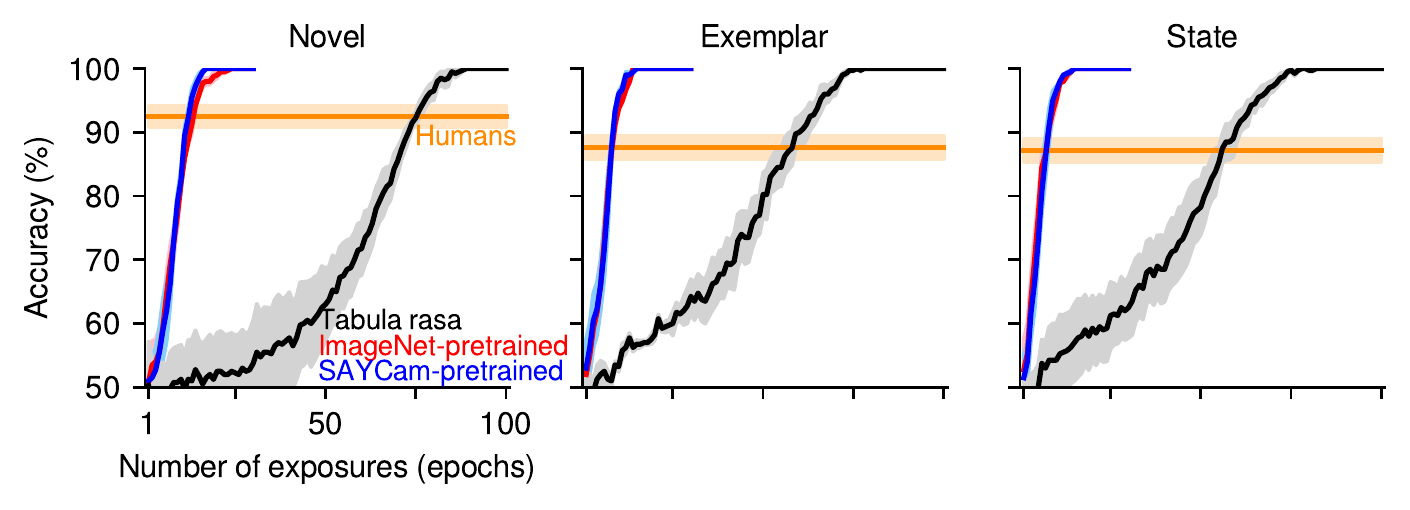}
  \caption{Recognition accuracy of the models in the \textit{novel}, \textit{exemplar}, and \textit{state} conditions, respectively. Human performance is indicated by the orange horizontal lines. Note that humans achieved this performance after just a single exposure to the study items during the study phase. The lines indicating human performance were continued along the \textit{x}-axis for visualization purposes only. For humans, error bars represent standard errors across participants; for models, error bars are standard errors over four different runs of the simulated experiment.}
  \label{brady2008_result_fig}
\end{figure}

Figure~\ref{brady2008_result_fig} shows our main results, where we have plotted the test accuracy of the models in the \textit{novel}, \textit{exemplar}, and \textit{state} conditions as a function of the number of exposures to the study stimuli. Tabula rasa models require dozens of passes through the study set in order to reach human-level accuracy on the recognition memory tests (recall that humans need only a single exposure to the study stimuli to reach this level of accuracy). 

Prior knowledge induced via pretraining improves the recognition memory performance of the models tremendously, however this improvement is not very visible after only a single exposure, so that even the pretrained models are roughly at chance level after a single exposure and they typically still need $\sim$10 exposures to the study stimuli in order to reach human-level recognition memory performance. 

We did not observe a significant difference between the performance of the ImageNet-pretrained vs.~SAYCam-pretrained models in recognition memory tests (red vs.~blue lines in Figure~\ref{brady2008_result_fig}). This result is surprising given the differences in the size and nature of these two datasets and points to a relatively generic, content non-specific benefit of pretraining for recognition memory performance, as will be discussed in more detail below.

In the human experiments of \citet{brady2008}, participants performed slightly better in the novel condition than in the exemplar and state conditions (horizontal orange lines in Figure~\ref{brady2008_result_fig}) Intriguingly, we observed a reversal of this pattern for the models such that they typically performed better in the exemplar and state conditions than in the novel condition (\textit{e.g.}, after 10 exposures to the study stimuli, the SAYCam-pretrained model achieved a test accuracy of $82.5\pm1.75\%$ (s.e.m.) in the novel condition, $96\pm0.5\%$ in the exemplar condition, and $96.5\pm0.75\%$ in the state condition). It remains unclear at the moment what causes this discrepancy between the human and the model results.

\subsection{Effects of model size, pretraining data size, and stimulus type on recognition memory performance}
Recent work in deep learning has revealed scale (both model size and training data size) to be a critical determinant of model capabilities, with increases in scale sometimes leading to surprising qualitative jumps in model capabilities \citep{brown2020,ramasesh2021,ramesh2022,chowdhery2022,kaplan2020,hoffmann2022}. Motivated by these recent findings, we conducted experiments to investigate the effects of model size, pretraining data size, and stimulus type on the memory efficiency of our iGPT models in the visual recognition memory experiment of \citet{brady2008}.

\begin{figure}
  \centering
    \includegraphics[width=1.0\textwidth, trim=0mm 0mm 0mm 0mm, clip]{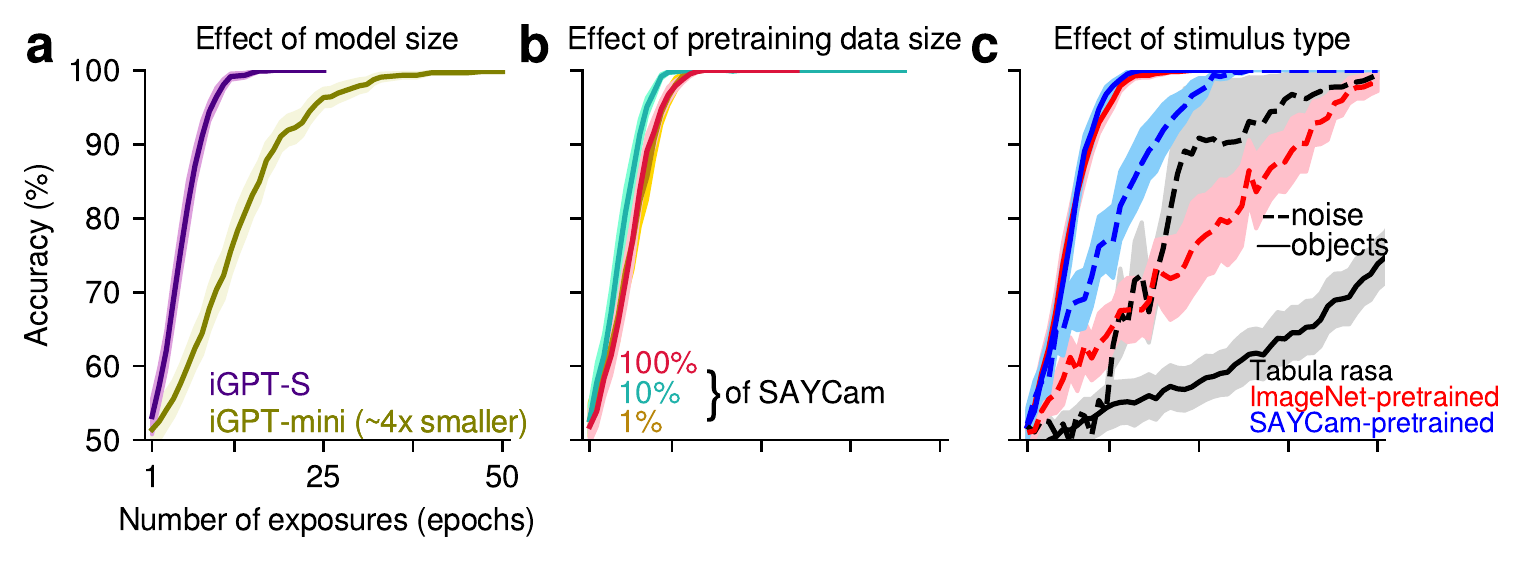}
  \caption{The effect of model size (a), pretraining data size (b), and stimulus type (c) on the recognition accuracy of the models. For these results, we average over different test conditions to come up with a single recognition accuracy score. Error bars represent standard errors over four different runs of the simulated experiments.}
  \label{scaling_fig}
\end{figure}

\textbf{Effect of model size:} Figure~\ref{scaling_fig}a shows the effect of model size on recognition memory accuracy for the ImageNet-pretrained models (in Figure~\ref{scaling_fig}, we show the recognition accuracy averaged over the three test conditions). In this experiment, we first pretrained both an iGPT-S model and a $4\times$ smaller iGPT-mini model on ImageNet up to the same loss value ($\sim$1.5 nats/pixel) and then performed the recognition memory experiment of \citet{brady2008} with both models. Model size had a clear effect on recognition memory accuracy: even after controlling for the pretraining loss, hence controlling for the amount of prior knowledge in the models, the larger model (iGPT-S) performed significantly better than the 4$\times$ smaller model (iGPT-mini). However, this effect was again not very visible after a single exposure. The difference between the models rather became apparent only in later exposures. 

After a single exposure to the study items, the iGPT-S model had a recognition accuracy of $53.2\pm2.4\%$ and the iGPT-mini model had a recognition accuracy of $51.4\pm2.5\%$. Thus, a quick back-of-the-envelope calculation suggests that assuming every quadrupling of the model size improves single-exposure recognition accuracy by $\sim$2\%, it would take roughly 18 quadruplings of the model size, or roughly 11 orders of magnitude increase in model size, to reach human-level recognition accuracy, starting from an iGPT-S model. Since the iGPT-S model already has $O(10^8)$ parameters, this means that a model with $O(10^{19})$ parameters would be needed to reach human-level recognition accuracy.\footnote{Alternatively, assuming a $3\%$ improvement per quadrupling of model size leads to an estimate of $O(10^{15})$ parameters and a $4\%$ improvement per quadrupling leads to an estimate of $O(10^{13})$ parameters needed to reach human-level recognition memory performance.~For comparison, GPT-3 has $O(10^{11})$ parameters and the human brain has an estimated $O(10^{15})$ synapses.} This corresponds to a model with roughly four orders of magnitude more parameters than the number of synapses in the human brain (although see the Caveats section below for some caveats about this quick back-of-the-envelope estimate). 

\textbf{Effect of pretraining data size.} Figure~\ref{scaling_fig}b shows the effect of pretraining data size on recognition memory accuracy. In this experiment, we first pretrained iGPT-S models on 100\%, 10\%, and 1\% of the SAYCam dataset and then performed the recognition memory experiments of \citet{brady2008} on these models. The pretraining data size had very little to no effect on the recognition memory performance. Together with the insensitivity of recognition memory performance to pretraining data type (ImageNet vs.~SAYCam), this again points to a relatively generic benefit for pretraining in these experiments.  

\textbf{Effect of stimulus type.} Finally, we also investigated the effect of stimulus type to be remembered on recognition memory performance. We compared the recognition memory performance with object images used in \citet{brady2008} to recognition memory performance with simple random \textit{iid} noise stimuli (see the Appendix for stimulus details). The results are shown in Figure~\ref{scaling_fig}c. For tabula rasa models (black lines), recognition memory performance with noise stimuli (dashed lines) was actually better than the performance with object images. This is presumably because the noise stimuli are intrinsically more distinct from each other, whereas the object images are intrinsically more similar. However, the pattern was reversed for models pretrained on natural images. Both for the ImageNet-pretrained model and for the SAYCam-pretrained model, recognition performance was better with object images than with noise stimuli. This makes sense, as pretraining with natural images presumably biases the model to spend more of its capacity toward natural images, hence improves the encoding of those kinds of images at the expense of other types of possible images (such as random noise). 

\subsection{No evidence for a conceptual distinctiveness advantage for recognition memory in the models}

\begin{figure}
  \centering
    \includegraphics[width=1.0\textwidth, trim=0mm 0mm 0mm 0mm, clip]{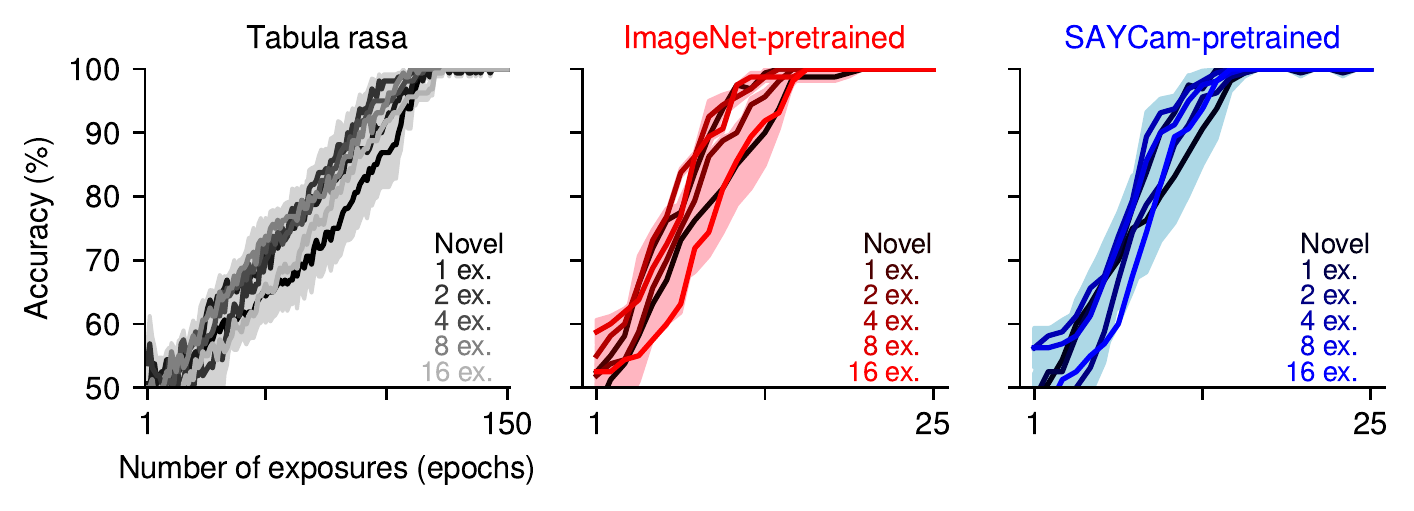}
  \caption{Results from the simulated \citet{konkle2010} experiment probing the effect of conceptual distinctiveness on recognition memory accuracy. Results are shown for both tabula rasa, ImageNet-pretrained, and SAYCam-pretrained models. In each subplot, different lines represent different test conditions corresponding to different conceptual distinctiveness levels (more exemplars from a category make those exemplars conceptually less distinctive). There is no clear and consistent effect of conceptual distinctiveness on recognition memory accuracy for either of the three models.}
  \label{konkle_fig}
\end{figure}

In a follow-up study to \citet{brady2008}, \citet{konkle2010} set out to investigate some of the factors behind the large capacity of human visual recognition memory. They focused, in particular, on the role of conceptual distinctiveness of the study items (\textit{i.e.} whether the items belong to the same or different categories) in affecting the human recognition memory performance. To this end, they conducted a study similar to that of \citet{brady2008}, but this time they carefully controlled the number of distinct exemplars from different object categories during the study phase. Some categories had 16 distinct exemplars presented during the study phase, some had 8, some 4, some 2, and some had only 1 exemplar. In the test trials, they observed that conceptual distinctiveness had a significant effect on the recognition accuracy of the subjects: when the study item and the foil belonged to different categories (\textit{novel} condition), recognition accuracy was at 93\%; accuracy decreased as the study item and the foil both belonged to increasingly more ``cluttered'' categories: \textit{e.g.} when they both belonged to a 16-exemplar category (\textit{16 exemplar} condition), recognition accuracy dropped to 83\%. \citet{konkle2010} argued that this result was specifically due to the conceptual or categorical distinctiveness of the items and could not be attributed to perceptual distinctiveness (perceptual distinctiveness was a confounding factor in their experiment). 

We replicated the experiment of \citet{konkle2010} with our iGPT models (see the Appendix for further experimental details) and the results are shown in Figure~\ref{konkle_fig}.~Unlike in the human experiment, we did not find a clear and consistent effect of conceptual distinctiveness on recognition memory accuracy. This result suggests that our iGPT models, including the models pretrained on natural images, likely encode these object images quite differently from the human subjects, in a way that is less reliant on ``conceptual hooks''.

\section{Caveats}
Here, we would like to mention two important limitations/caveats that should be borne in mind in interpreting the results presented in this work. Future work should address these limitations:
\begin{itemize}
    \item We have only considered a particular model class, namely Image GPT, for modeling the image likelihoods $p_\theta(\mathbf{x})$ in this work. Future work should consider other types of likelihood-based generative models, such as vector-quantized variational autoencoders \citep{vandenoord2017,razavi2019}, vector-quantized generative adversarial networks \citep{esser2021}, or flow-based generative models \citep{kingma2018}, to make sure the results in this work are not overly sensitive to the choice of model class. The same caveat applies to hyperparameter choices/configurations not explored in this paper.
    \item Although we have shown that model size and pretraining data size do not have a significant effect on the recognition memory accuracy of the models after only a single exposure to the study items (Figure~\ref{scaling_fig}), it is conceivable that for even larger model sizes or larger pretraining data sizes than those tested here the models could display a qualitative jump in single-exposure performance, significantly outperforming all the models presented here and possibly even approaching human recognition memory performance. Future work should consider this possibility by testing even larger model sizes and larger pretraining data sizes than those that were feasible for us to implement in this work.
\end{itemize}

\section{Discussion}
Recent work in deep learning found that large language models like GPT-2 can ``unintentionally'' memorize large amounts of \textit{verbatim} text from their training data \citep{carlini2019,carlini2021}, which would imply a large recognition memory capacity for such models (but with unknown \textit{efficiency}), however a direct quantitative comparison with humans in a domain where human recognition memory is particularly strong, \textit{i.e.} in the visual domain, has not been performed before. To our knowledge, this paper presents the first attempt to carry out such a rigorous, quantitative comparison.

It would be informative to perform similar comparisons in other (non-visual) domains. It is known, for instance, that auditory recognition memory is substantially inferior to visual recognition memory in humans \citep{cohen2009}. It would be interesting to know if this result holds for gradient descent trained deep learning models too. Similarly, some classic studies exist in the human memory literature on recognition memory for linguistic materials, \textit{i.e.}~words and sentences \citep{shepard1967}.~It would be interesting to compare these results with recognition memory for similar linguistic stimuli in large language models.

Our results suggest that current gradient-based deep learning algorithms may be inadequate for achieving human-level efficiency in incorporating new visual information into long-term memory. In practical terms, we can imagine broadly two types of responses to these findings, one more dismissive and the other more constructive. The dismissive response might go as follows:
\begin{itemize}
    \item Humans and deep learning models operate under very different constraints, hence human-level speed or efficiency to incorporate new information into memory may not be critically important for deep learning models. For example, in standard machine learning problems, data is often stored externally to the model, so one can always do another ``pass'' over the data and with a sufficiently large capacity model, one would be guaranteed to learn it perfectly with enough passes over it. 
\end{itemize}
Although this response contains some kernel of truth, it is ultimately not satisfactory. First, not all problems in machine learning involve externally stored data: \textit{e.g.}~real-time, streaming data sources create conditions and constraints a lot closer to those under which the human brain evolved to operate. Secondly, even in more standard machine learning problem settings, it would be of great practical interest to improve the efficiency of our models to incorporate new information (inspired by the superior human efficiency to do the same). Most obviously, this would, for example, significantly reduce the training time of our models.

The constructive response would consider our results as a challenge to algorithmically improve the memory efficiency of deep learning models. Several ideas have already been proposed in the literature to achieve similar goals:
\begin{itemize}
    \item Adding fast-learning Hebbian terms to parameter updates as in fast weights \citep{hinton1987}, activity caching \citep{rae2018}, or weight imprinting \citep{qi2018}. However, these ideas either lack the generality of gradient descent or have not been shown to scale up to very large problems. It is important to demonstrate that such ideas do not limit the power of gradient-based learning in large scale problems.
    \vspace{1em}
    \item Using large explicit memory stores to cache compressed representations of the observed data or a subset thereof \citep{grave2016,blundell2016,orhan2018,khandelwal2019,borgeaud2021,wu2021}. This proposal has its drawbacks: for example, it creates very large memory storage and demanding retrieval requirements at inference time. Nevertheless, it would be an effective way to solve the fast learning problem considered in this paper (perhaps even trivial depending on the particular implementation adopted). Something like this may also be part of the solution animal brains found at evolutionary timescale to deal with the fast learning problem, an idea known as the \textit{complementary learning systems} hypothesis in psychology and neuroscience \citep{mcclelland1995,kumaran2016}.
\end{itemize}


In this work, we have not addressed the important question of the long-term retention of new information (or its durability in memory), but only its induction or incorporation into a model (or into long-term memory). There are some classic results in the human memory literature on the long-term retention of memories in humans \citep{bahrick1975,bahrick1984}. A rigorous, quantitative comparison of the retention capabilities of humans vs.~deep learning models trained via gradient descent would be highly informative as well.

In this work, we have only considered recognition memory, but recognition is only a single facet of memory in humans and machine learning models. Future work should consider other (more demanding) aspects of memory such as recall and the use of retrieved information in downstream tasks.

Finally, even when it comes to recognition memory, we have only compared single-exposure recognition memory in humans vs.~gradient-based deep learning models in this work. It is important to note that the results may be different for multiple exposures. In humans, experimental evidence suggests that visual recognition memory improves over multiple exposures to the same objects \citep{miner2020}, just like in deep learning models. However, whether humans would still outperform deep learning models for a given number of exposures in a specific problem setting remains an open question.

\section*{Acknowledgements}
I am grateful to the authors of \citet{brady2008} and \citet{konkle2010} for sharing their experimental stimuli. I am grateful to the late Jorge Luis Borges for having written \textit{Funes the Memorious}, which was a direct inspiration for this work. I am grateful to the HPC team at NYU for doing such an excellent job of maintaining the computing cluster. It is needless to say all errors in this work, conceptual errors, errors of execution, errors of omission, and more, are mine and mine only: ``I should have liked to produce a better paper, but the time in which I could improve it is now gone.''

\bibliography{brady}
\bibliographystyle{apalike}

\section*{Appendix}
\subsection*{Training details}
The ImageNet- and SAYCam-pretrained models were both trained for 15 epochs on the ImageNet and SAYCam datasets respectively, as described in the main text. We were able to bring the training loss down to $\sim$1.45 nats/pixel on ImageNet and down to $\sim$1.15 nats/pixel on SAYCam after 15 epochs of training (here ``pixel'' refers to a pixel in the reduced image space). Figure~\ref{unconditional_samples_fig} below shows some unconditional samples generated from these ImageNet- and SAYCam-pretrained models.

In all experiments reported in this paper, we used the Adam optimizer with a learning rate of 0.0005 and a batch size of 32. Larger batch sizes were computationally not feasible for us due to the large memory requirements of the iGPT-S model. Both pretraining on ImageNet and SAYCam as well as training on the study sets of \citet{brady2008} and \citet{konkle2010} were implemented with data parallelism over 32 NVIDIA RTX 8000 GPUs (with 48 GB GPU memory) or over 16 A100 GPUs (with 80 GB GPU memory).

With the capacity of the iGPT-S model, it was possible to achieve near-zero training loss values for training datasets with thousands of images, given enough passes over the dataset. Figure~\ref{losses_on_study_fig} below shows example training loss curves on the study set of \citet{brady2008} (with 2500 training images) for the tabula rasa, ImageNet-pretrained, and SAYCam-pretrained iGPT-S models.

\subsection*{Experimental details}
For the simulated \citet{brady2008} and \citet{konkle2010} experiments, we generated two different versions of the study and test sets and ran experiments with each version using two different random seeds, yielding a total of four different runs of each experiment. In the simulated \citet{konkle2010} experiments, there were 40 categories each in the 16-exemplar, 8-exemplar, 4-exemplar, 2-exemplar, and 1-exemplar categories in the study set, for a total of 40*(16+8+4+2+1)=1240 study stimuli. The test session consisted of 40 trials each of the \textit{novel}, \textit{16-exemplar}, \textit{8-exemplar}, \textit{4-exemplar}, \textit{2-exemplar}, and \textit{1-exemplar} conditions for a total of 240 test trials.

For the random noise experiments reported in Figure~\ref{scaling_fig}c in the main text, we generated random 64$\times$64 images, where the value of each pixel was uniformly and independently sampled from one of 512 discrete values (identical in size to the reduced natural images used elsewhere in this work). We generated 2500 such random images for the study set and generated a separate set of 300 novel random images that were used in the recognition test trials (for a total of 300 test trials). We repeated each random noise experiment four times with different random seeds.

\begin{figure}
  \centering
    \includegraphics[width=1.0\textwidth, trim=0mm 0mm 0mm 0mm, clip]{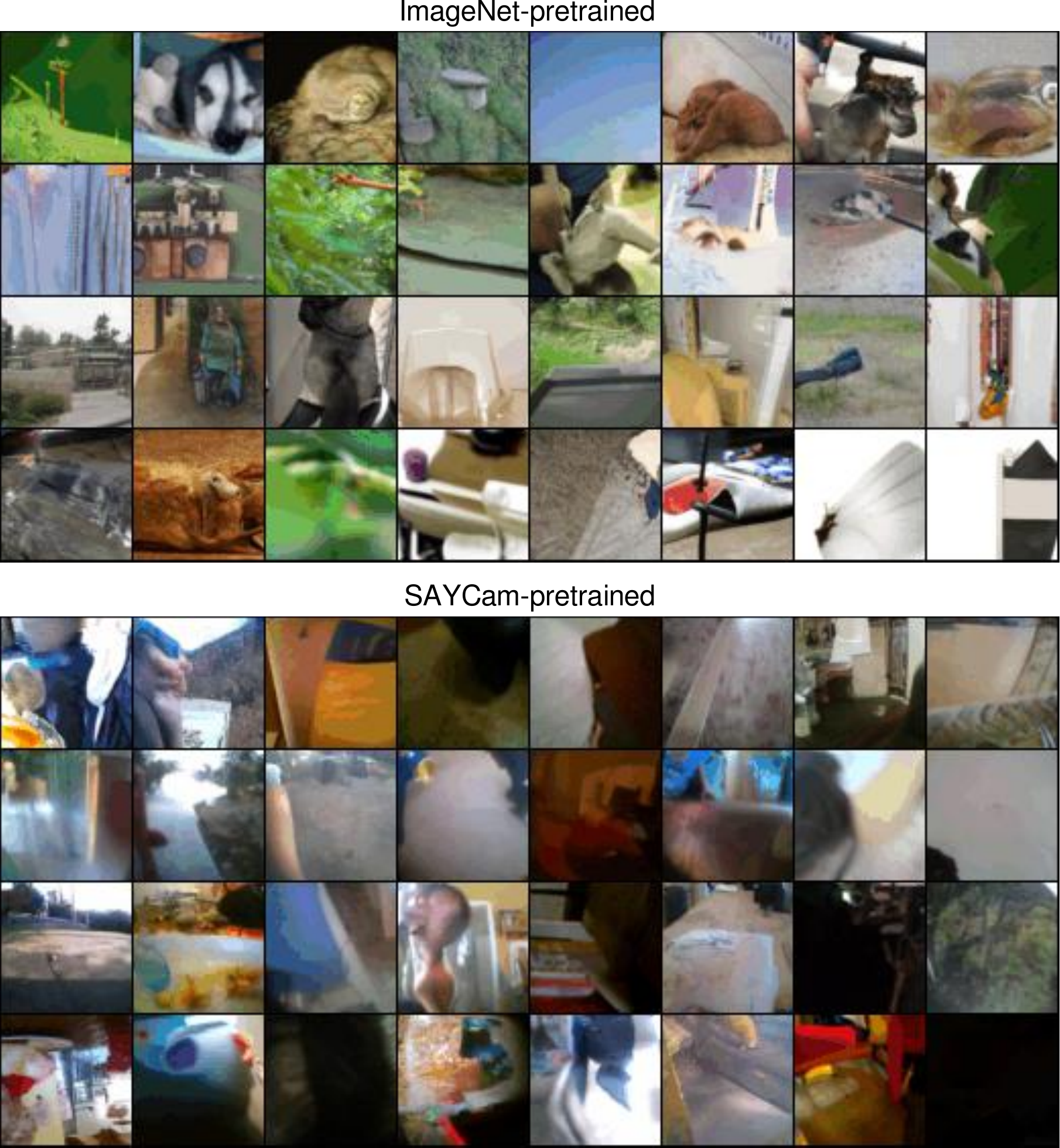}
  \caption{Unconditional samples generated from the ImageNet-pretrained and SAYCam-pretrained iGPT-S models.}
  \label{unconditional_samples_fig}
\end{figure}

\begin{figure}
  \centering
    \includegraphics[width=0.8\textwidth, trim=0mm 0mm 0mm 0mm, clip]{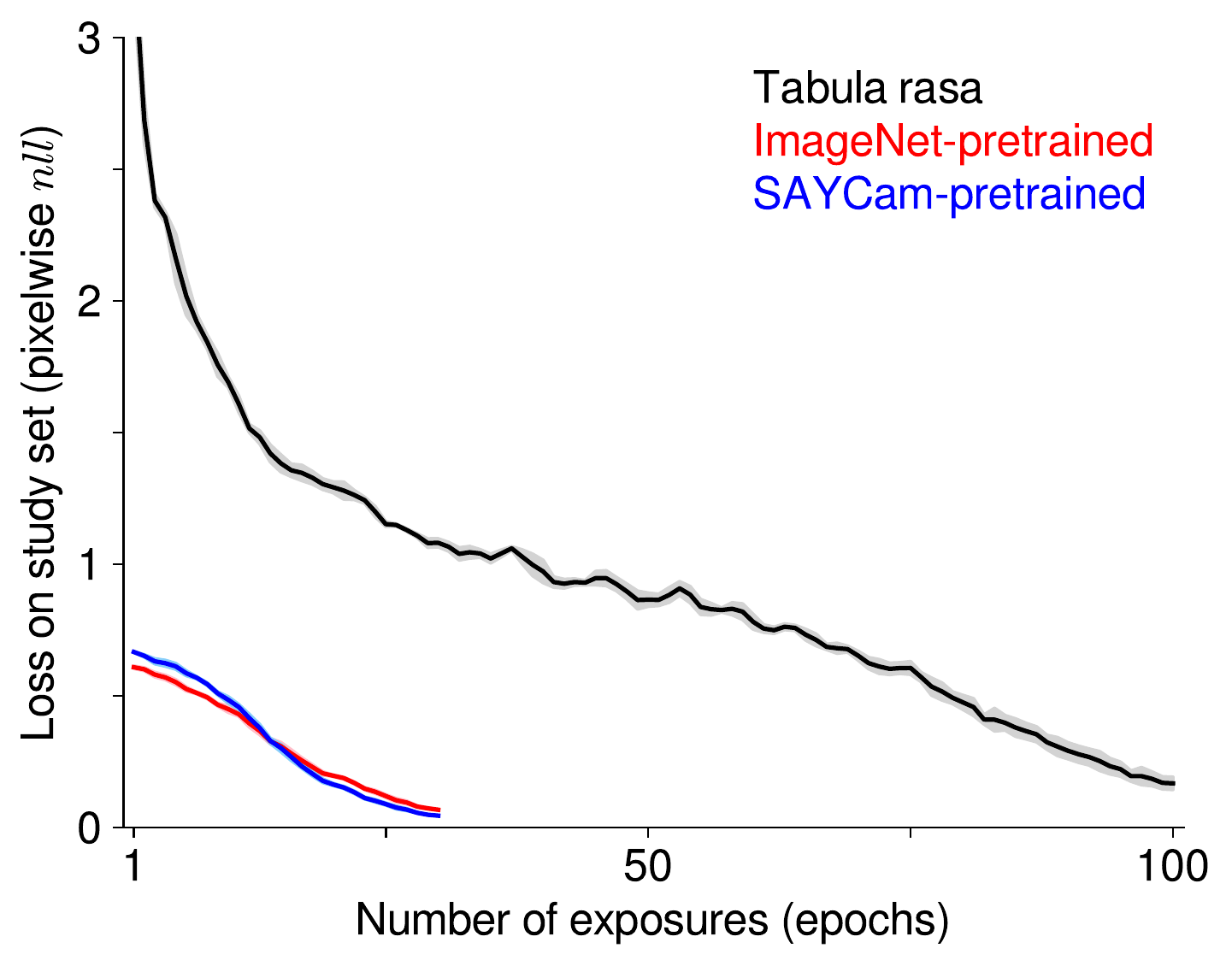}
  \caption{Training losses on the study set of \citet{brady2008} (with 2500 training images) for the tabula rasa, ImageNet-pretrained, and SAYCam-pretrained iGPT-S models. For all three models, it is possible to achieve near-zero training loss given enough passes over the training images. Error bars represent standard errors over four different runs of the experiment. \textit{nll}: negative log-likelihood.}
  \label{losses_on_study_fig}
\end{figure}

\end{document}